\documentclass[lettersize,journal]{IEEEtran}
\usepackage{amsmath,amsfonts,amssymb}
\usepackage{algorithmic}
\usepackage{algorithm}
\usepackage{array}
\usepackage[caption=false,font=normalsize,labelfont=sf,textfont=sf]{subfig}
\usepackage{textcomp}
\usepackage{stfloats}
\usepackage{url}
\usepackage{verbatim}
\usepackage{graphicx}
\usepackage{cite}
\usepackage[colorlinks,
            linkcolor=blue,
            anchorcolor=blue,
            citecolor=blue
            ]{hyperref}

\usepackage{booktabs,flushend}
\usepackage{nameref}
\usepackage{orcidlink}

\begin{document}
\title{Learning the Basis: A Kolmogorov-Arnold Network\\ Approach Embedding Green's Function Prior}

\author{Rui Zhu,~\IEEEmembership{Student~Member,~IEEE,}
        Yuexing Peng,~\IEEEmembership{Member,~IEEE,}\\
        George C. Alexandropoulos,~\IEEEmembership{Senior Member,~IEEE}, Peng Wang\,~\IEEEmembership{Member,~IEEE},\\ Wenbo Wang\,~\IEEEmembership{Senior Member,~IEEE}, and Wei Xiang\,~\IEEEmembership{Senior Member,~IEEE}

\thanks{R. Zhu, Y. Peng, and W. Wang are with the Key Laboratory of Universal Wireless Communication, Ministry of Education, School of Information and Communication Engineering, Beijing University of Posts
and Telecommunications, Beijing 100876, China (e-mails: \{rayyyy, yxpeng\}@bupt.edu.cn).}
\thanks{G. C. Alexandropoulos is with the Department of Informatics and Telecommunications, National and Kapodistrian University of Athens, Panepistimiopolis Ilissia, 16122 Athens, Greece (e-mail: alexandg@di.uoa.gr).}
\thanks{P. Wang is with a telecommunication company in Stockholm, Sweden (e-mail: wp\_ady@hotmail.com).} 
\thanks{W. Xiang is with the School of Computing, Engineering and Mathematical Sciences, La Trobe University, Melbourne, VIC 3086, Australia (e-mail: w.xiang@latrobe.edu.au).}}


\maketitle
\begin{abstract}
This letter proposes PhyKAN, a novel electromagnetics-informed learning framework that embeds the governing Electric Field Integral Equation (EFIE) directly into a dual-branch architecture for surface-current modeling. Inspired by the Kolmogorov-Arnold representation theorem, PhyKAN formulates a physics-informed framework where the local branch learns geometry-adaptive basis functions via Kolmogorov–Arnold Networks (KAN), while the global branch incorporates Green’s function priors to preserve long-range interactions and physical consistency. This design offers a unified formulation that generalizes classical Rao-Wilton-Glisson (RWG) basis into a data-driven, adaptive, and interpretable representation. It is demonstrated that, across canonical geometries, PhyKAN achieves reconstruction relative errors below 3\%, preserves accurate RCS patterns, and remains robust across ablation, frequency, and mesh-resolution tests, with interpretable KAN-based complexity.
\end{abstract}


\section{Introduction}

\IEEEPARstart{S}{olving} electromagnetic (EM) scattering via the Electric Field Integral Equation (EFIE) is central to computational electromagnetics (CEM)~\cite{ref1}, with the Method of Moments (MoM) being the standard approach. This method expands the induced surface current using predefined basis functions, typically the Rao–Wilton–Glisson (RWG) basis~\cite{ref2}. Despite its success, MoM remains fundamentally non-learnable because both the mesh-prescribed basis space and the Green’s function operator kernel~\cite{ref16} are fixed, limiting its ability to adapt to geometric or physical complexity.


To address the latter limitation, several directions have been explored. Higher-order basis functions enrich MoM expressiveness using higher-degree polynomials~\cite{ref4,ref5,ref6}, while characteristic or macro-basis constructions improve efficiency through physics-informed domain decomposition~\cite{ref7,ref8}. On another front, instead of improving the basis itself, Adaptive Mesh Refinement (AMR) refines the discretization in regions of high error~\cite{ref9,ref10}. Although effective, these approaches remain hand-crafted and problem-specific, and they retain the fixed discretization structure of MoM, leaving the overall formulation fundamentally non-adaptive.

Deep learning techniques are recently emerging as efficient tools in CEM problems, mainly along two directions: black-box current or field prediction without any explicit basis representation~\cite{refA,refB}, and MoM acceleration through surrogate models, operator learning, or iterative refinement, still relying on fixed basis functions~\cite{refC,refD,refE}. More recent studies integrate physical structure into neural models, such as EFIE-guided graph learning~\cite{Stylianopoulos,2508}. This indicates that neither direction addresses the non-learnable nature of the basis functions, which remain unable to adapt to EM physics.

Recent progress in scientific machine learning enables neural network architectures that integrate physical structure withing learnable representations. The Kolmogorov–Arnold Network (KAN)~\cite{ref11} is one of the well suited learning frameworks for this purpose that replaces fixed activations with learnable univariate functions~\cite{ref12}, providing adaptable basis behavior. KAN is flexible in both kernel choices~\cite{ref13,ref14} and graph-based extensions~\cite{ref15} that embed geometric and physical interactions, making it a strong foundation for a physics-guided learnable basis for the MoM approach.


In this letter, we introduce PhyKAN, a dual-branch, physics-informed framework that embeds the EFIE directly into a learnable architecture.
The key innovation is to replace the fixed Rao–Wilton–Glisson basis with KAN-based adaptive basis functions, allowing the surface-current representation to follow geometric and physical variations.
Meanwhile, the global branch incorporates the Green’s function kernel as an explicit prior, lifting it into a higher-dimensional latent space to model long-range coupling while preserving EFIE consistency.
By uniting KAN-driven basis adaptation with a Green’s-function-embedded global prior, PhyKAN offers a compact, physically grounded reformulation of EFIE–MoM within a modern neural architecture.

\section{Surface Current as a Learning Problem}


Let $\mathbf{G}(\cdot) \in \mathbb{C}^{3\times 3}$ denote the dyadic Green's function in three-dimensional free space, $\boldsymbol{J}(\cdot)\in \mathbb{C}^{3\times 1}$ the induced surface current, and $\boldsymbol{E}_{\rm inc}(\cdot)\in \mathbb{C}^{3\times 1}$ the incident field. Let $\boldsymbol{n}(\boldsymbol{p})$ denote the outward unit normal vector at $\boldsymbol{p}$ on the PEC surface $\mathcal{S}$. By applying the boundary conditions of a Perfectly Electrically Conducting (PEC) object with surface~$\mathcal{S}$, the EFIE can be formulated $\forall \boldsymbol{p} \in \mathcal{S}$ as:

\begin{equation}
\label{jf}
\mathbf{n}(\boldsymbol{p}) \times\left[\mathbf{E}_{\text {inc }}(\boldsymbol{p})+\int_{\mathcal{S}} \mathbf{G}\left(\boldsymbol{p}, \boldsymbol{p}^{\prime}\right) \boldsymbol{J}\left(\boldsymbol{p}^{\prime}\right) \mathrm{d}\mathcal{S}^{\prime}\right]=\mathbf{0}, \quad \boldsymbol{p} \in \mathcal{S}
\end{equation}


Here, $\mathbf{G}\left(\boldsymbol{p}, \boldsymbol{p}^{\prime}\right)$ governs the electromagnetic interaction, while $\boldsymbol{J}\left(\boldsymbol{p}^{\prime}\right)$ is expanded through MoM basis functions. Together, these two elements define the core structure that PhyKAN reformulates into a learnable and physically consistent representation.
%
\subsection{Learning the Basis}
The MoM discretizes this integral equation~\eqref{jf} by expanding the unknown current $\boldsymbol{J}(\boldsymbol{p})$ using a set of predefined basis functions. Typically, the RWG basis $\{{f}_n(\cdot)\}_{n=1}^3$ comprising fixed and geometry-defined functions~\cite{ref2} is used, yielding the following surface current approximation:
\begin{equation}
\label{rwg}
\boldsymbol{J}\left(\boldsymbol{p}_{k}'\right) \approx \sum_{n=1}^3 I_{n,k} {f}_n\left(\boldsymbol{p}_{k}'\right),
\end{equation}
where $I_{n,k}$ are the unknown complex coefficients of the $n$-th RWG basis function on the $k$-th triangular element centered at $\boldsymbol{p}_{k}'$. This geometry-fixed basis restricts the MoM’s adaptability to complex current distributions. 

In the classical MoM formulation, all degrees of freedom reside in the scalar coefficients $\{I_{n,k}\}$, whereas the basis shapes $\{f_n(\cdot)\}$ are entirely dictated by the mesh geometry and remain identical across different illuminations or target configurations. The RWG family is edge-based and divergence-conforming, which makes it physically sound but also rigid: improving expressiveness typically relies on mesh refinement or higher-order/macrobasis constructions, increasing the number of unknowns rather than enriching the basis itself.




To introduce adaptability while preserving the MoM structure, we reinterpret basis construction itself as a learnable function approximation problem. Rather than only solving for fixed coefficients $I_{n,k}$ in~\eqref{rwg} that scale a geometry-defined basis, we seek an optimal, data-driven basis family $\{b_n(\cdot;\boldsymbol{u})\}$ whose functional shape can adapt to the local geometry and electromagnetic response, while the expansion in~\eqref{learnable_basis} keeps the familiar MoM form, yielding the following generalized expansion:
\begin{equation}
\boldsymbol{J}(\boldsymbol{p}'_{k}) \approx \sum^{N_b}_{n=1} \alpha_{n,k}b_n(\boldsymbol{p}'_{k};\boldsymbol{u}).
\label{learnable_basis}
\end{equation}
Here, $N_{b}$ denotes the number of learnable basis functions used per element, $\alpha_{n,k}$ is the coefficient associated with the $n$-th learnable basis on the $k$-th triangular element, and $b_n(\cdot;\boldsymbol{u})$ is a learnable function parameterized by $\boldsymbol{u}$, whose goal is to adapt its functional shape locally based on the underlying physics and geometry. Notably, setting $b_{n} = f_n$ and $\alpha_{n,k} = I_{n,k}$ recovers the classical MoM expansion in~\eqref{rwg} as a special case. 

This formulation is enabled by the Kolmogorov–Arnold representation theorem~\cite{ref17}, which expresses multivariate functions as compositions of univariate nonlinear mappings. Motivated by this structure, each basis function is parameterized using a KAN representation:




\begin{equation}
\label{KAN_basis_expansion}
b_n\left(\boldsymbol{p}'_{k} ; \boldsymbol{u}_n\right)=\sum_{i=1}^M c_{n, i} \phi\left(\boldsymbol{w}_{n, i}^{\top} \boldsymbol{p}_{k}'+\beta_{n, i}\right),
\end{equation}
where $\phi(\cdot)$ plays the same role as the univariate nonlinear mappings $\phi_{ij}(\cdot)$ in~\cite{ref17}, and $\left\{\boldsymbol{w}_{n, i},\beta_{n, i},c_{n, i}\right\}$ are learnable parameters controlling each kernel’s orientation, offset, and amplitude, respectively. The term $\phi\left(\boldsymbol{w}_{n, i}^{\top} \boldsymbol{p}_{k}'+\beta_{n, i}\right)$ performs a learnable linear projection of the spatial coordinate $\boldsymbol{p}'_{k}$, allowing each kernel to adapt to different geometric directions and field variations.

Substituting (\ref{KAN_basis_expansion}) into (\ref{learnable_basis}), yields the approximation:
\begin{equation}
\label{KAN_MoM_form}
\boldsymbol{J}(\boldsymbol{p}'_{k}) \approx \sum_{n=1}^{N_b} \sum_{i=1}^M
\tilde c_{n,i,k} \phi\left(\boldsymbol{w}_{n, i}^{\top} \boldsymbol{p}_{k}'+\beta_{n, i}\right),
\end{equation}
with $\tilde{c}_{n, i, k} \triangleq \alpha_{n, k} c_{n, i}$ acting as the learnable counterpart of each $I_{n,k}$ and $\phi\left(\boldsymbol{w}_{n, i}^{\top} \boldsymbol{p}_{k}'+\beta_{n, i}\right)$'s serving as data-driven, adaptive generalizations of the fixed RWG basis functions. The resulting expansion mirrors the MoM form in \eqref{rwg} while replacing the static RWG basis with a KAN-based, physics-adaptive representation.
\subsection{Embedding Green’s Function Physics}

The dyadic Green’s function $\mathbf{G}\left(\boldsymbol{p}, \boldsymbol{p}^{\prime}\right)$ in~\eqref{jf} fully characterizes the electromagnetic interaction kernel in the EFIE and governs long-range coupling across the surface. In classical MoM, this operator is fixed and directly used to assemble the system matrix, leaving no room for learning enriched representations of global physics. To introduce learnability without violating the EFIE structure, PhyKAN incorporates $\mathbf{G}$ as an explicit physics prior and then lifts this operator into a higher-dimensional latent space. Formally, for the $k$-th element we construct
\begin{equation}
\mathbf{F}_k=\Phi\left(\mathbf{G}\left(\boldsymbol{p}_k, \boldsymbol{p}_{k^{\prime}}\right)\right)
\end{equation}
where $\Phi(\cdot)$ is a linear mapping initialized to satisfy $\Phi(\mathbf{G})=\mathbf{G}$, ensuring exact physical fidelity at the start of training.  During optimization, $\Phi$ lifts the Green’s kernel $\mathbf{G}$ into a latent representation $\mathbf{F}_k \in \mathbb{C}^D$($D>3$) preserving the EFIE prior while enabling enriched, geometry-aware representations of long-range coupling.

This high-dimensional lifting contrasts sharply with classical MoM, where the Green’s kernel is confined to a fixed $3 \times3$ dyadic structure. By embedding $\mathbf{G}$ as a prior and expanding its representational space, PhyKAN retains strict physical consistency while unlocking additional flexibility for capturing complex, nonlocal interactions.

\section{Proposed PhyKAN Framework}

\begin{figure*}[!t]
\centering
\includegraphics[width=18cm]{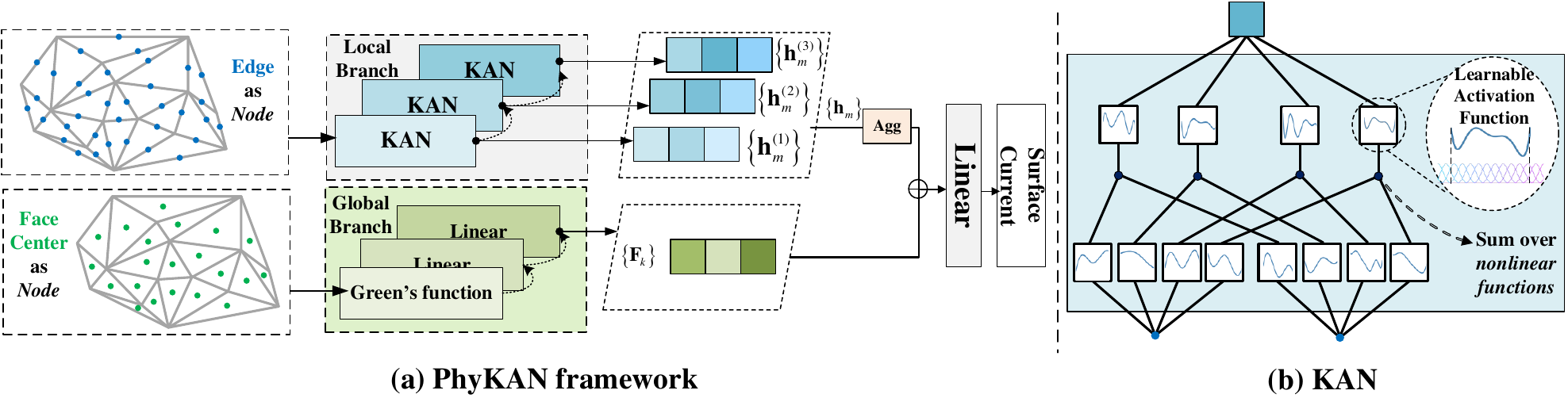}
\caption{(a) PhyKAN consists of a KAN-based local branch that encodes geometry-adaptive basis features and a global branch that incorporates Green’s function priors for physically consistent modeling of long-range electromagnetic interactions. (b) The conventional KAN structure~\cite{ref11}, where learnable activation functions form adaptive univariate kernels.}

\label{fig_2}
\end{figure*}

In this section, we build upon the preceding formulation into PhyKAN, an EFIE-derived, physics-guided learning framework.
The proposed artificial neural network, whose architecture is visualized in Fig.~\ref{fig_2}(a), has a dual-branch structure that enables adaptive basis learning by separately modeling local geometric features and global EM interactions.


\subsection{Neural Network Architecture}

The \textbf{local branch} learns edge-wise basis representations by treating each RWG basis function as a node, consistent with the MoM formulation in which each RWG function is defined on the shared edge of two triangular faces. The input is a set of $M$ edge-nodes, each described by a geometric feature vector $\mathbf{g}_m \in \mathbb{R}^d$ where the dimension $d$ of the feature vector includes edge length, adjacent-face normals, and relative edge orientation. As illustrated in Fig.~\ref{fig_2}(a), the geometric descriptor is passed through a three-layer KAN encoder. Each KAN layer performs a nonlinear lifting of $\mathbf{g}_m$ through spline-based learnable activations:
\begin{equation}
\mathbf{h}^{(l)}_m\triangleq\mathrm{KAN}_{\boldsymbol{u}}\left(\mathbf{g}_m\right),
\end{equation}
where $\mathbf{h}^{(l)}_m \in \mathbb{R}^D$ denotes the $l$-th layer embedding. Each layer contains learnable linear projections and kernel activations (B-spline, rational, or Fourier), enabling the basis to adapt its local functional shape.

To obtain a stable and expressive embedding, the three KAN outputs are fused via a residual linear aggregator:
\begin{equation}
\mathbf{h}_m=\sum_{l=1}^3 {\eta}_l \odot \mathbf{h}_m^{(l)},
\end{equation}
where ${\eta}_l$ are learnable per-layer weighting vectors that fuse multi-level KAN features into the final geometry-adaptive edge embedding.

The local feature $\mathbf{H}_k$ of each $k$-th triangular element is aggregated from its three associated edge local embeddings of the corresponding edges ($\mathbf{h}_a$, $\mathbf{h}_b$, and $\mathbf{h}_c$), as follows:
\begin{equation}
\mathbf{H}_k=\frac{1}{3}\left(\mathbf{h}_a+\mathbf{h}_b+\mathbf{h}_c\right),
\end{equation}
providing a learnable counterpart of the RWG basis local expansion.

The \textbf{global branch} of PhyKAN models long-range electromagnetic interactions across the surface. In contrast to the local branch where each node corresponds to a mesh edge, here one element-node is defined per triangular element. The input to this branch is a set of $K$ element-nodes, where each node $k$ is represented by the centroid $\boldsymbol{p}_k \in \mathbb{R}^3$ of the $k$-th element, together with the incident direction $(\theta,\phi)$ of the scattered EM fields $\boldsymbol{E}_{\rm inc}(\boldsymbol{p}_{k})$'s. To embed electromagnetic physics into this branch, the first-layer coupling weights between element-nodes are initialized directly with the dyadic Green’s function:
\begin{equation}
\mathbf{W}_{k,j}^{(0)}=\mathbf{G}\left(\boldsymbol{p}_{k}, \boldsymbol{p}_{j}\right),
\end{equation}
which encodes the physical interaction between elements $k$ and $j$ as governed by the EFIE kernel. This initialization ensures that global message passing is grounded in the true nonlocal behavior of Maxwell’s equations. 

To incorporate the dependence on incidence, the incident angles $(\theta, \phi)$ are mapped into a $D_{\rm ang}$-dimensional (with $D_{\rm ang}\triangleq 4L$) Fourier embedding, as follows:
\begin{equation}
\label{eq:fourier_embed}
\boldsymbol{\gamma}(\theta,\phi) =
\big[
\sin(\omega_1 \theta), \cos(\omega_1 \theta), \ldots,
\sin(\omega_L \phi), \cos(\omega_L \phi)
\big],
\end{equation}
where $\{\omega_\ell\}_{\ell=1}^{L}$ are frequency parameters.
In the sequel, the global coupling features and the latter $D_{\rm ang}$-directional embedding are concatenated and sequentially passed through two linear projections, yielding:
\begin{equation}
\mathbf{F}_k=\mathbf{W}^{(2)}\left(\sigma\left(\mathbf{W}^{(1)}\left(\sum_{j=1}^K \mathbf{W}_{k j}^{(0)} \boldsymbol{p}_j\right) \oplus \boldsymbol{\gamma}(\theta, \phi)\right)\right)+\mathbf{b},
\end{equation}
where $\oplus$ denotes concatenation, $\sigma(\cdot)$ is an activation function, and $\mathbf{W}^{(1)}$, $\mathbf{W}^{(2)}$, and $\mathbf{b}$ are trainable parameters.

To finally predict surface currents, the local basis feature $\mathbf{H}_k$ and the global physical context $\mathbf{F}_k$ are combined as follows:
\begin{equation}
\mathbf{z}_k = \mathbf{H}_k \oplus \mathbf{F}_k,
\end{equation}
and are then projected to the $6$-dimensional output $\boldsymbol{J}_{\rm pre}(\boldsymbol{p}_k)\in \mathbb{C}^{3\times 1}$ via a linear head.

\subsection{Physics-Informed Loss Function}
For each incident field $\boldsymbol{E}^{(m)}_{\text{inc}}$, we compute the EFIE residual over all $K$ surface elements, rather than supervising the currents with MoM solutions.
The residual for the $m$-th incident field is written as:
\begin{equation}
\mathcal{L}^{(m)} 
= \frac{1}{K} \sum_{k=1}^{K}
\left\|
\boldsymbol{n}(\boldsymbol{p}_k) \times 
\left(
    \boldsymbol{E}^{(m)}_{\text{inc}}(\boldsymbol{p}_k)
    + \sum_{j=1}^{K}
      \boldsymbol{G}(\boldsymbol{p}_k,\boldsymbol{p}_j)
      \boldsymbol{J}^{(m)}_{\text{pre}}(\boldsymbol{p}_j)
\right)
\right\|_2^2
\end{equation}
where $\boldsymbol{p}_k$, $\boldsymbol{p}_j$  denote the centers of the centers of the $k$-th and $j$-th mesh elements. This residual enforces the predicted current $\boldsymbol{J}^{(m)}_{\text{pre}}$ to satisfy the PEC boundary condition for the incident field $\boldsymbol{E}^{(m)}_{\text{inc}}$. 
The overall physics-informed loss is obtained by averaging over a batch of $N_s$ incident fields:
\begin{equation}
\mathcal{L}_{\text{EFIE}}=\frac{1}{N_s} \sum_{m=1}^{N_s} \mathcal{L}^{(m)}.
\end{equation}
This formulation enables PhyKAN to be trained without MoM current labels, relying purely on the discretized EFIE for physical supervision.

\section{Numerical Results and Discussion}


To evaluate PhyKAN, we assess its effectiveness, physical consistency, and adaptive behavior as a learnable basis. High-fidelity MoM solutions on a fine mesh ($\lambda/10$), the standard CEM reference~\cite{ref1}, serve as ground truth for both surface-current reconstruction and Radar Cross Section (RCS) validation. The model takes triangular mesh elements with EM descriptors and predicts induced currents. Experiments cover four canonical PEC targets~\cite{2508} (cube, sphere, cone, assembly body) and a complex SLICY object~\cite{slicy} under a 1 GHz, unit-amplitude, vertically polarized plane wave with $\theta,\phi\in[0^\circ,180^\circ]$.
DL-based CEM baselines either black-box the current mapping or only accelerate MoM without learnable bases; hence, we benchmark PhyKAN variants under a unified setting against MoM.



\begin{figure*}[!t]
\centering
\includegraphics[width=18cm]{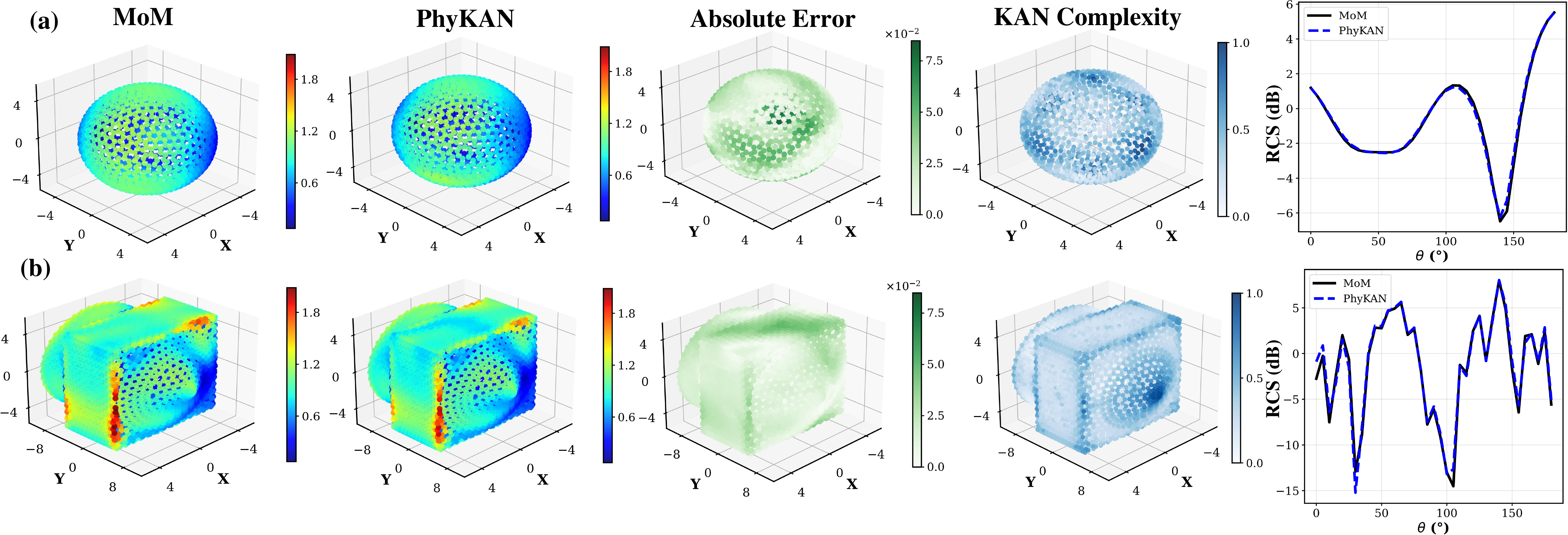}
\caption{Surface current reconstruction on (a) the sphere and (b) the assembly body. Columns show the MoM reference, PhyKAN prediction, absolute error, KAN complexity maps, and the corresponding RCS comparison.}
\label{fig_3}
\end{figure*}

\begin{table}
\centering
\caption{Relative MSE (×$10^{-2}$) of PhyKAN and ablated variants.}
\renewcommand{\arraystretch}{1.4} 
\begin{tabular}{c|c|c|c|c}
\hline
\textbf{Method} & \textbf{Cube} & \textbf{Sphere} & \textbf{Cone} & \textbf{Assembly Body} \\ \hline

PhyKAN \textit{(no physics init)}   & 3.42 & 2.95 & 3.88 & 6.73 \\ 

PhyKAN \textit{(MLP Basis)}   & 4.75 & 3.89 & 5.26 & 8.94 \\ 

\textbf{PhyKAN}    & 1.21 & 0.94 & 1.28 & 1.84 \\

 \hline
\end{tabular}
\label{table: da}
\end{table} 

\begin{table}
\centering
\caption{Relative MSE (×$10^{-2}$) of PhyKAN at different frequencies.}
\renewcommand{\arraystretch}{1.4}
\begin{tabular}{c|c|c|c|c}
\hline
\textbf{Frequency} & \textbf{Cube} & \textbf{Sphere} & \textbf{Cone} & \textbf{Assembly} \\ \hline
1 GHz & 1.21 & 0.94 & 1.28 & 1.84 \\
2 GHz & 1.46 & 1.07 & 1.53 & 2.21 \\
5 GHz & 1.89 & 1.33 & 1.98 & 2.67 \\ \hline
\end{tabular}
\label{table: freq}
\end{table}




Figure~\ref{fig_3} shows the reconstruction results.
Across both the sphere and the assembly body, PhyKAN closely matches the MoM current distributions, with absolute errors remaining small and localized. 
The KAN complexity maps quantify the average activation magnitude of all kernels in KAN blocks across the three edges of each element, yielding higher values near geometric discontinuities and thus highlighting PhyKAN’s adaptive allocation of $b_n(\cdot;\boldsymbol{u})$ compared with the fixed RWG basis.
The far-field RCS results indicate that PhyKAN matches MoM patterns despite receiving no RCS supervision.

Table~\ref{table: da} reports the relative MSE for the ablation study. Removing the Green’s function initialization leads to higher errors across all geometries, while replacing the KAN basis with a plain MLP causes an even larger degradation. The full PhyKAN model achieves the best accuracy, demonstrating the complementary benefits of physics-based initialization and the learnable KAN basis.
Table~\ref{table: freq} summarizes PhyKAN’s accuracy when trained at 1, 2, and 5~GHz.
The error increases only slightly with frequency due to finer high-frequency current oscillations, and remains stable across all targets, with a modest 1.3–1.6× rise from 1 to 5~GHz.

\begin{figure}[!t]
\centering
\includegraphics[width=9cm]{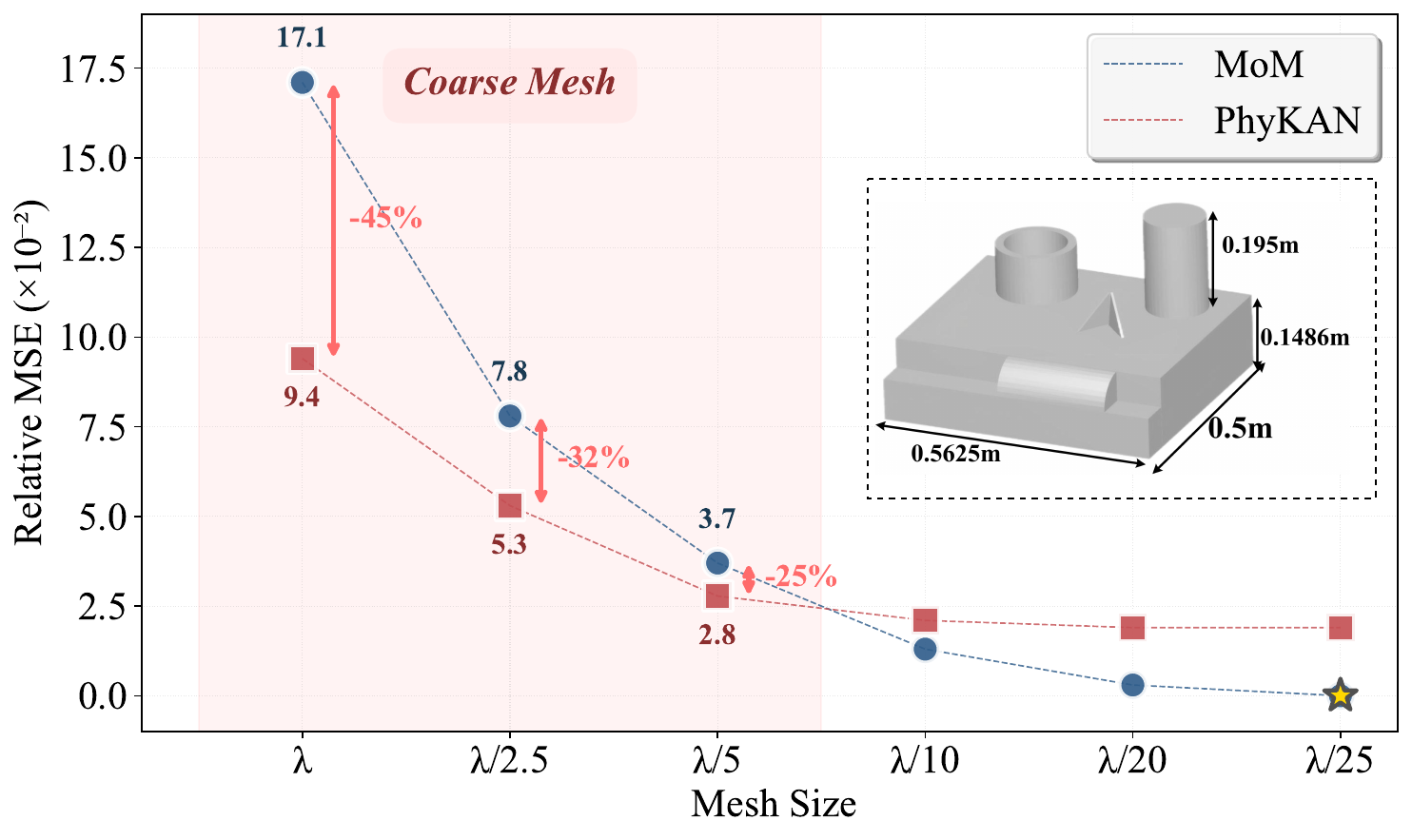}
\caption{Relative MSE (×$10^{-2}$) of PhyKAN and MoM under different mesh resolutions, computed with respect to the $\lambda$/25 MoM reference.}
\label{fig_5}
\end{figure}

Fig.~\ref{fig_5} evaluates performance under various mesh resolutions, using relative MSE with $\lambda/25$ MoM result as reference. In coarse-mesh regimes (e.g., $\lambda$, $\lambda/2.5$, $\lambda/5$), PhyKAN substantially reduces MSE on coarse meshes by compensating for geometric undersampling with its learnable basis. MoM overtakes only at very fine resolutions where computational cost dominates, making PhyKAN most valuable in coarse-mesh regimes.

\section{Conclusion}

This letter introduced PhyKAN, an EFIE-derived framework that replaces the fixed RWG basis with an adaptive learnable representation while embedding the Green’s function as a physics prior, forming a unified and physically consistent architecture for electromagnetic modeling. Future work will focus on scaling the learnable basis to large-scale and broadband scenarios.

\vfill


\newpage

\begin{thebibliography}{1}
\bibliographystyle{IEEEtran}

\bibitem{ref1}
S. M. Rao, D. R. Wilton, and A. W. Glisson, “Electromagnetic scattering by surfaces of arbitrary shape,” \emph{IEEE Trans. Antennas Propag.}, vol. 30, no. 3, pp. 409–418, May 1982.

\bibitem{ref2}
R. F. Harrington and J. L. Harrington, \emph{Field Computation by Moment Methods}. Oxford, U.K.: Oxford Univ. Press, Inc., 1996.


\bibitem{ref4}
G. Kang, J. Song, W. C. Chew, K. C. Donepudi, and J.-M. Jin, “A novel grid-robust higher order vector basis function for the method of moments,” \emph{IEEE Trans. Antennas Propag.}, vol. 49, no. 6, pp. 908–915, Jun. 2001.

\bibitem{ref5}
O. S. Kim and P. Meincke, “Adaptive integral method for higher order method of moments,” \emph{IEEE Trans. Antennas Propag.}, vol. 56, no. 8, pp. 2298–2305, Aug. 2008.

\bibitem{ref6}
N. Zhang, Y. Chen, Y. Ren, and J. Hu, “A modified HODLR solver based on higher order basis functions for solving electromagnetic scattering problems,” \emph{IEEE Antennas Wireless Propag. Lett.}, vol. 21, no. 12, pp. 2452–2456, Dec. 2022.

\bibitem{ref7}
E. Lucente, A. Monorchio, and R. Mittra, “An iteration-free MoM approach based on excitation independent characteristic basis functions for solving large multiscale electromagnetic scattering problems,” \emph{IEEE Trans. Antennas Propag.}, vol. 56, no. 4, pp. 999–1007, Apr. 2008.

\bibitem{ref8}
P. Du, Z.-W. Tong, H. Lin, and G. Zheng, “Wideband RCS analysis of finite periodic array in the vicinity of object using modified SSED-CBFM and improved FIR,” \emph{IEEE Antennas Wireless Propag. Lett.}, vol. 24, no. 6, pp. 1447–1451, Jun. 2025.


\bibitem{ref9}
T. Plewa, T. Linde, and V. G. Weirs, Eds., \emph{Adaptive Mesh Refinement: Theory and Applications, Proc. Chicago Workshop on Adaptive Mesh Refinement Methods, Sep. 3–5, 2003}. Berlin, Germany: Springer, 2005.



\bibitem{ref10}
A. Amor-Martin and L. E. Garcia-Castillo, “Adaptive semi-structured mesh refinement techniques for the finite element method,” \emph{Appl. Sci.}, vol. 11, no. 8, p. 3683, Apr. 2021.

\bibitem{refA}
J. H. Kim and S. W. Choi, “A deep learning-based approach for radiation pattern synthesis of an array antenna,” \emph{IEEE Access}, vol. 8, pp. 226059–226063, 2020.

\bibitem{refB}
M. Salucci, M. Arrebola, T. Shan, and M. Li, “Artificial intelligence: New frontiers in real-time inverse scattering and electromagnetic imaging,” \emph{IEEE Trans. Antennas Propag.}, vol. 70, no. 8, pp. 6349–6364, Aug. 2022.



\bibitem{refC}
M. Baldan, P. Di Barba and D. A. Lowther, "Physics-informed neural networks for inverse electromagnetic problems," \emph{IEEE Trans. Magn.}, vol. 59, no. 5, pp. 1-5, May 2023.

\newpage

\bibitem{refD}
C. Brennan and K. McGuinness, “Site-specific deep learning path loss models based on the method of moments,” \emph{Proc. 17th Eur. Conf. Antennas Propag. (EuCAP)}, Florence, Italy, 2023. 



\bibitem{refE}
D. -H. Kong, J.-N. Cao, W.-W. Zhang, W.-C. Huang, X.-Y. He and L. Liu, "An AI predictor: From point clouds to scattered far fields for 3-D PEC targets," \emph{IEEE Trans. Antennas Propag.}, vol. 72, no. 6, pp. 5179-5190, Jun. 2024.



\bibitem{Stylianopoulos}
K. Stylianopoulos, P. Gavriilidis, G. Gradoni, and G. C. Alexandropoulos, “Graph-CNNs for RF imaging: Learning the electric field integral equations,” in \textit{Proc. European Signal Process. Conf. (EUSIPCO)}, Palermo, Italy, Sep. 2025, pp. 1-5.


\bibitem{2508}
R. Zhu, Y. Peng, P. Wang, G. C. Alexandropoulos, W. Wang, and W. Xiang, 
“U-PINet: End-to-end hierarchical physics-informed learning with sparse graph coupling for 3D EM scattering modeling,” 
\textit{arXiv preprint arXiv:2508.03774}, Aug. 2025.

\bibitem{ref11}
Z. Liu, Y. Wang, S. Vaidya, F. Ruehle, J. Halverson, M. Soljačić, T. Y. Hou, and M. Tegmark, “KAN: Kolmogorov–Arnold networks,” in \textit{Proc. 13th Int. Conf. Learn. Represent. (ICLR)}, Singapore, Apr. 2025, pp. 1346-1354.


\bibitem{ref12}
J. D. Toscano, V. Oommen, A. J. Varghese, Z. Zou, N. A. Daryakenari, C. Wu, and G. E. Karniadakis, “From PINNs to PIKANs: Recent advances in physics-informed machine learning,” \textit{Mach. Learn. Comput. Sci. Eng.}, vol. 1, no. 15, Mar. 2025. 


\bibitem{ref13}
J. Xu, Z. Chen, J. Li, S. Yang, W. Wang, X. Hu, and E. C. H. Ngai, “FourierKAN-GCF: Fourier Kolmogorov-Arnold network– An effective and efficient feature transformation for graph collaborative filtering,” \emph{CoRR}, vol. abs/2406.01034, 2024.


\bibitem{ref14}
X. Yang and X. Wang, “Kolmogorov–Arnold Transformer,” in \emph{Proc. Int. Conf. Learn. Represent. (ICLR)}, Singapore, Apr. 2025, pp. 76063–76086.


\bibitem{ref15}
L. Li, Y. Zhang, G. Wang, and K. Xia, “Kolmogorov–Arnold graph neural networks for molecular property prediction,” \textit{Nat. Mach. Intell.}, vol. 7, pp. 1346–1354, Aug. 2025.


\bibitem{ref16}
G. Barton, \emph{Elements of Green’s Functions and Propagation: Potentials, Diffusion, and Waves}. Oxford, U.K.: Clarendon Press, 1989, p. 465.

\bibitem{ref17}

A. N. Kolmogorov, “On the representation of continuous functions of several variables as superpositions of continuous functions of a smaller number of variables,” \textit{Dokl. Akad. Nauk SSSR}, vol. 108, no. 2, pp. 179–182, 1956.


\bibitem{ref18}
M.~A.~Richards, \emph{Fundamentals of Radar Signal Processing}, 2nd ed. New York, NY, USA: McGraw-Hill, 2014.

\bibitem{slicy}
L. M. Yuan, Y. G. Xu, W. Gao, et al., “Design of scale model of plate-shaped absorber in a wide frequency range,” \emph{Chin. Phys. B}, vol. 27, no. 4, p. 044101, 2018.


\end{thebibliography}
\end{document}